\pdfoutput=1
\documentclass[10pt,twocolumn,letterpaper]{article}

\usepackage[pagenumbers]{cvpr}
\usepackage{times}
\usepackage{epsfig}
\usepackage{graphicx}
\usepackage{amsmath}
\usepackage{amssymb}
\usepackage{booktabs}
\usepackage{enumitem}

\begin{document}

\title{Decoupled Representation Learning for Face Presentation Attack Detection under Smart Glass Capture}

\author{Raghavendra Ramachandra$^1$ \ Narayan Vetrekar$^2$ \ Neha Naik$^2$ \ Rajendra Gad$^2$\\
$^1$ SAFE Centre, Norwegian University of Science and Technology (NTNU), Gj{\o}vik, Norway\\
$^2$ School of Physical and Applied Sciences, Goa University, Goa, India\\
{\tt\small raghavendra.ramachandra @ ntnu.no, vetrekarnarayan @ unigoa.ac.in }
}

\maketitle
\thispagestyle{empty}
{\let\thefootnote\relax\footnotetext{Preprint of a paper accepted at the International Joint Conference on Biometrics (IJCB 2026).}}

\begin{abstract}
Face presentation attack detection is typically studied under fixed camera or handheld acquisition, whereas smart glass capture introduces a more challenging setting with ego motion, changing viewpoint, background clutter, distance variation, and less controlled illumination. To address this challenge, we present \emph{SmaGla-FPAD}, a smart glass based face presentation attack database acquired using Ray-Ban Meta Wayfarer Smart Glasses across indoor, outdoor, and in-car settings. The database contains 297,000 sample images with both bona fide and silicone mask based attacks. We further propose \emph{DeCoupPAD-FAS++}, a decoupled representation learning framework that separates spoof related evidence from nuisance sensitive capture variation through complementary attack and style branches, pseudo-domain supervision, adversarial suppression, and prompt guided semantic alignment. Experiments on SmaGla-FPAD show that the proposed method achieves the best balance between attack rejection and bona fide acceptance in both in domain and cross scenario evaluation. Ablation results further confirm the contribution of disentanglement, pseudo-domain supervision, and prompt based regularisation under smart glass capture.
\end{abstract}

\section{Introduction}

Face recognition is widely deployed because it is contactless, convenient, and compatible with standard RGB cameras. It is used in applications ranging from border control and secure access to mobile authentication and remote identity verification. Although three dimensional and multispectral face recognition can offer stronger resilience to spoofing, their deployment is often limited by sensor cost and system complexity \cite{Survey_2024, Survey_2025, Ramachandra:2017:PAD:3058791.3038924}. In contrast, two dimensional face recognition remains dominant in practical systems because it can be implemented using conventional cameras and integrated readily into existing platforms \cite{DHS_PAD, Ramachandra:2017:PAD:3058791.3038924}. This is particularly relevant in mobile and wearable settings, where compact hardware and natural user interaction are essential.

\begin{figure}[t!]
	\centering
	\includegraphics[width=1\linewidth]{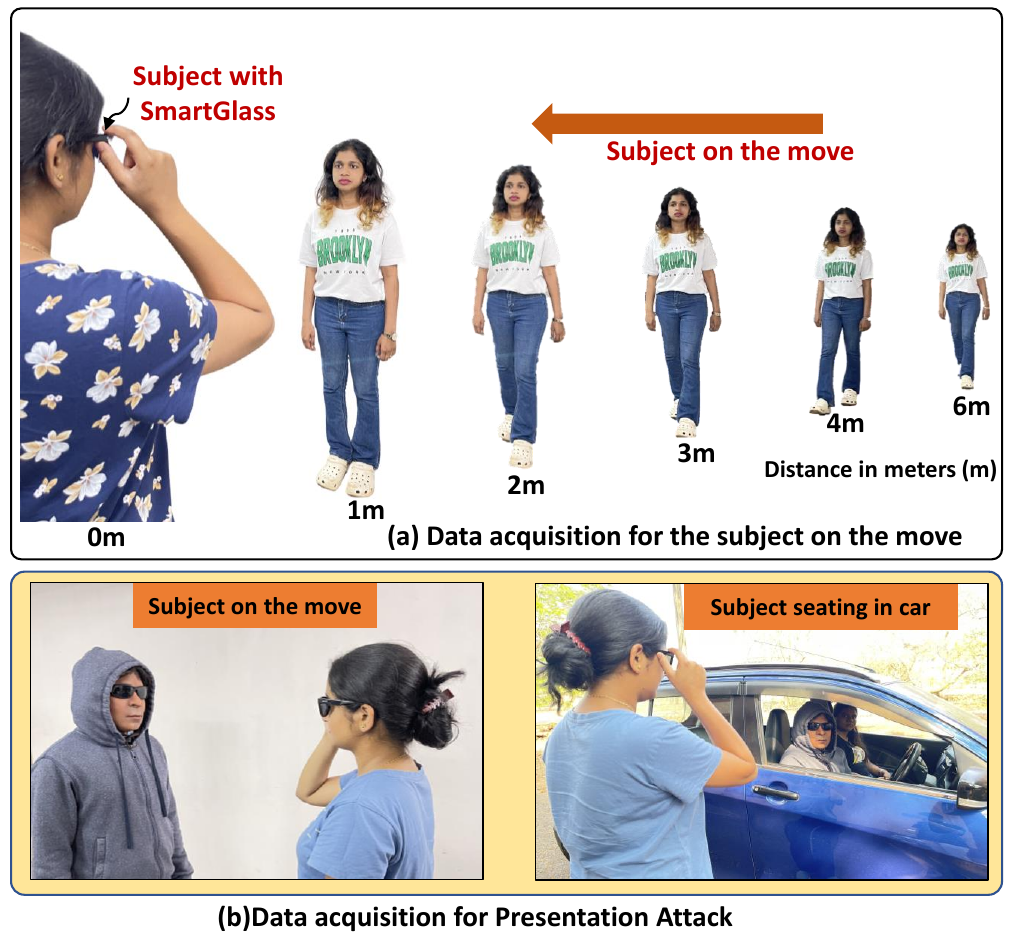}
	\caption{Illustration of the data acquisition scenario: (a) subject on the move towards the smart glass, (b) subject seated in the car.}
	\label{fig:data-acquisition}
\end{figure}

At the same time, RGB face recognition is vulnerable to presentation attacks, including printed photographs, display replays, photo wrapped media, silicone masks, and three dimensional facial replicas \cite{Ramachandra:2017:PAD:3058791.3038924, FaceMaskVuln, AVPAD_Survey, Survey_2024, Venkatesh-MADSurvey-IEEETTS-2021, DHS_PAD}. This threat is especially important in operational settings such as border control, where imaging conditions are less controlled and the capture process must remain efficient. Robust presentation attack detection (PAD) is therefore essential for trustworthy deployment.

Despite substantial progress in face PAD, generalisation remains a major challenge. Performance often degrades when illumination, motion, background, sensor placement, or attack presentation changes. This problem is more severe under smart glass capture, where the camera is body worn, the viewpoint is dynamic, and the recorded face is affected by motion blur, distance variation, background clutter, and unstable lighting. Existing PAD benchmarks rarely address this setting explicitly, and most prior methods are developed for fixed-camera or smartphone capture rather than first-person wearable acquisition. Consequently, face PAD still lacks both suitable datasets and models for robust smart glass capture.

In this work, we address this gap from both the data and modelling perspectives. First, we introduce \emph{SmaGla-FPAD}, a smart glass based face presentation attack database acquired using Ray-Ban Meta Wayfarer Smart Glasses in indoor, outdoor, and in-car settings. As shown in Fig.~\ref{fig:data-acquisition}, the dataset captures bona fide and silicone-mask-based attacks under realistic variation in scene, motion, and acquisition geometry. Second, we propose \emph{DeCoupPAD-FAS++}, a decoupled representation learning framework for face PAD under smart glass capture. The framework separates spoof discriminative evidence from nuisance sensitive capture variation through complementary attack and style branches, adversarial pseudo-domain suppression, and prompt guided image-text regularisation using a frozen vision language model.

Extensive experiments on the proposed dataset show that the method achieves the strongest overall trade-off between attack rejection and bona fide acceptance in both in-domain and cross-scenario evaluation. The results further indicate that robust smart glass PAD requires not only strong backbone features, but also explicit treatment of nuisance variation during training. The main contributions of this work are threefold:
\begin{itemize}[leftmargin=*,noitemsep, topsep=0pt,parsep=0pt,partopsep=0pt]
    \item We introduce \emph{SmaGla-FPAD}, a smart glass based face presentation attack database collected in indoor, outdoor, and in-car settings, with bona fide and silicone mask attack samples acquired under realistic operational variability.
    \item We propose \emph{DeCoupPAD-FAS++}, a decoupled representation learning framework that separates spoof related evidence from nuisance sensitive capture variation through complementary attack and style branches, pseudo-domain supervision, adversarial suppression, and prompt guided semantic alignment.
    \item We present in-domain, cross-scenario, and ablation studies showing that the proposed method provides the best  overall operating point under challenging smart glass capture conditions.
\end{itemize}
\section{Related Work}
\label{sec:related}
Face PAD has evolved from handcrafted descriptors to deep representation learning \cite{Survey_2025}. Early methods relied on texture, image quality, and motion cues. Local descriptors such as LBP and LBP TOP were used to capture fine surface differences between bona fide and spoof samples \cite{Raghusipco, ColorLBPfacePAD, BioEvaSpoof, Ramachandra:2017:PAD:3058791.3038924}, while quality based methods exploited artefacts such as specular reflection, blur, colour distortion, and sensor noise \cite{chang2022face_Qualitry, MSUMFSD_DB, raghavendra2015presentation}. Motion and physiological signal based approaches further incorporated cues such as remote photoplethysmography to separate live faces from print and replay attacks \cite{liu20163d, nowara2017ppgsecure, liu2018remote}. With the rise of deep learning, convolutional networks became the dominant paradigm for learning spoof related spatial and temporal features directly from RGB and depth data \cite{li2021diffusing, wang2022patchnet, qiao2022fgdnet, yang2014learn}. Subsequent work introduced attention mechanisms, transformer based architectures, anomaly detection, and one class formulations to capture broader facial context and improve robustness to unseen attacks \cite{huang2021multi, kong2022face, lucena2017transfer, george2021effectiveness_Super2, chen2019attention_Super3, wang2022face_Pix1, baweja2020anomaly, li2022one, george2020learning}. To address domain shift, several studies further incorporated domain adaptation and domain generalisation strategies based on adversarial alignment, distribution matching, and meta regularisation \cite{chen2021generalizable, shao2019multi, shao2020regularized, CF_FAS, cai2023rehearsal}.

More recent PAD research has increasingly explored foundation models, vision language learning, prompt based optimisation, and multimodal transformers. CLIP based and vision language approaches align visual spoof evidence with semantic prompt descriptions in order to improve cross domain generalisation and reduce dependence on dataset specific appearance statistics \cite{srivatsan2023flip, DomainPrompt, liu2024bottom, liu2024cfpl, liu2024fm, FLIP_FAS, ozgur2025foundpad, Transfer2025}. Prompt learning has been extended through knowledge guided prompt construction, gradient aware prompt tuning, and domain specific prompt design \cite{jiang2025learningPrompt, liu2025gradient, liu2024bottom}, while multimodal transformer models have exploited RGB, depth, and thermal cues through cross modal attention and expert routing \cite{liu2023fm, ma2025big}. Recent attack agnostic prompt formulations have also attempted to unify the treatment of physical and digital attacks \cite{li2025fa}. Despite these advances, most existing methods are developed for conventional fixed camera or handheld acquisition and do not explicitly address the strong nuisance variation introduced by smart glass capture. In contrast, the present work targets this setting directly and formulates smart glass PAD as a decoupled representation learning problem in which spoof evidence and capture dependent variation are modelled separately.
\begin{figure}[!]
	\centering
	\includegraphics[width=1\linewidth]{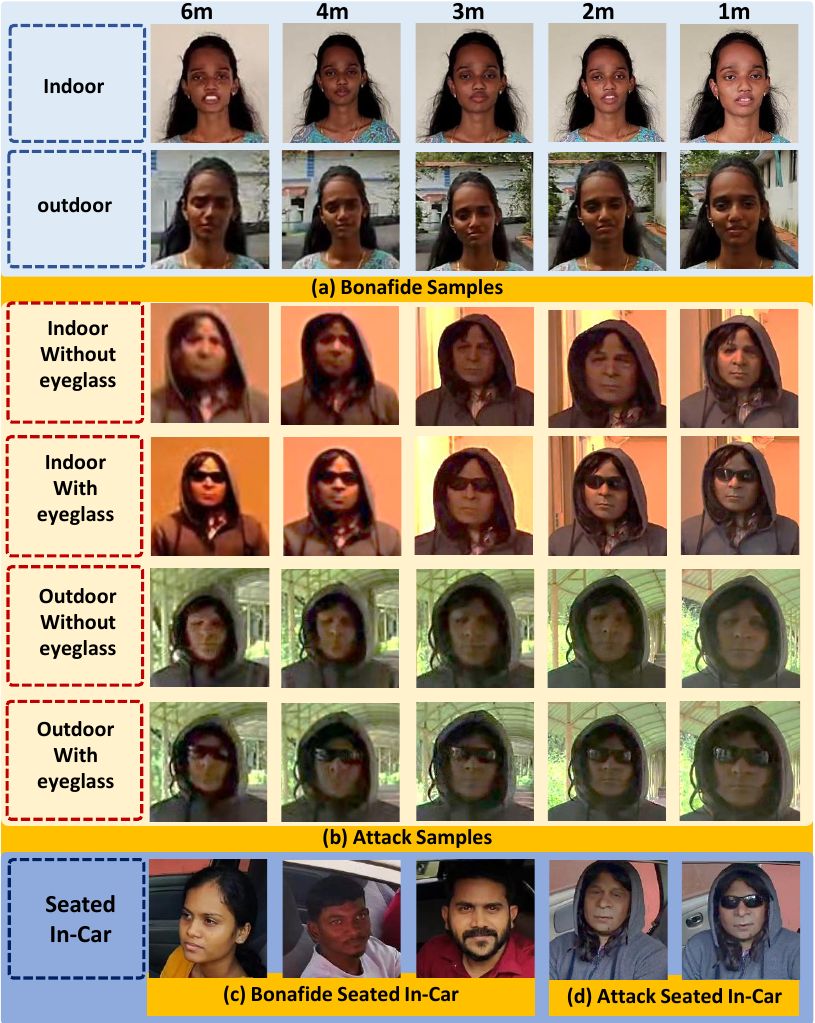}
	\caption{Sample images show the data acquired in SmaGla-FPAD in indoor, outdoor, and in-car scenarios.}
	\label{fig:sample-images}
\end{figure}
\begin{figure*}[htp]
	\centering
	\includegraphics[width = 0.95\linewidth]{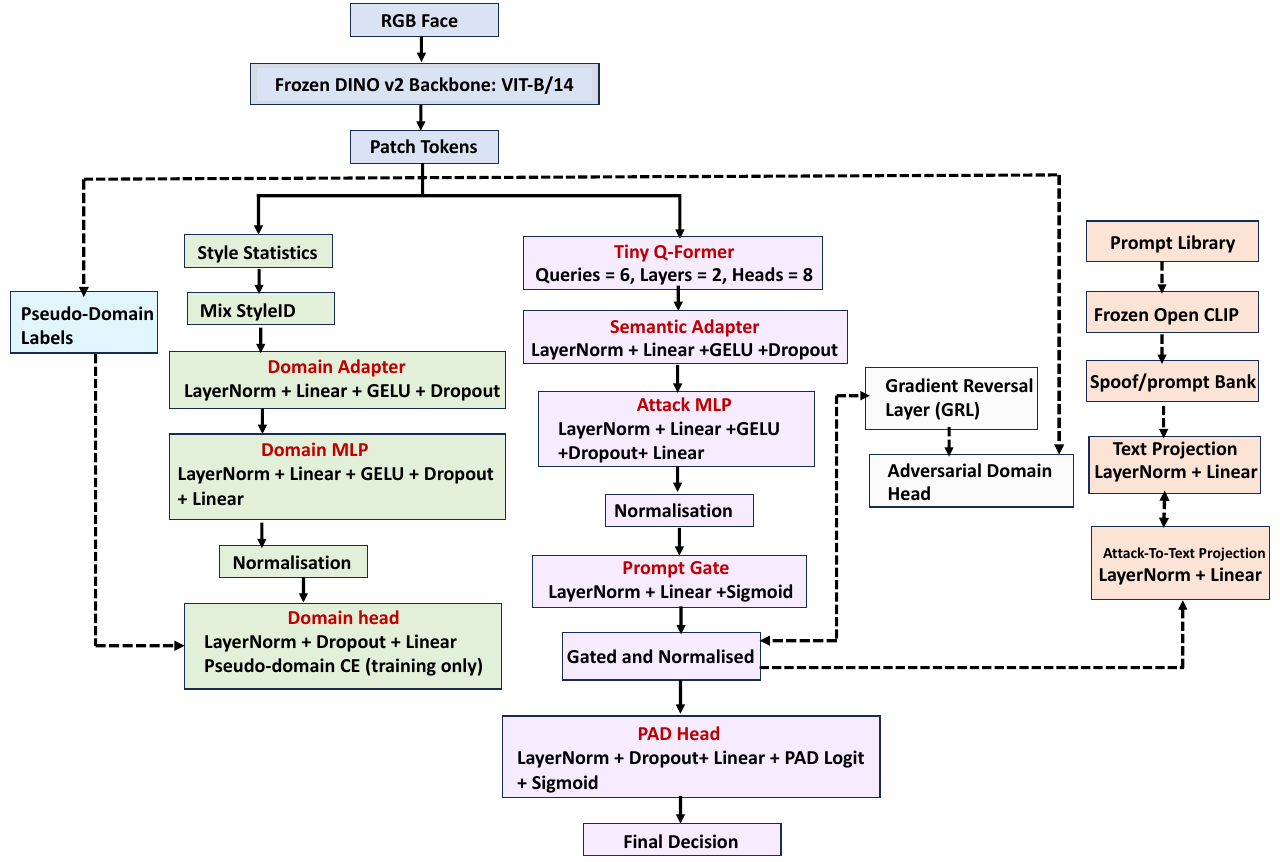}
	\caption{Overview of DeCoupPAD-FAS++. A frozen DINOv2 backbone extracts patch tokens that are decoupled into an attack content branch for PAD and a style domain branch for nuisance modelling. During training, the attack branch is regularised by adversarial pseudo-domain suppression and prompt guided image-text alignment using text projection and attack-to-text projection. Solid arrows denote inference, whereas dotted arrows indicate training only supervision.}
    \label{fig:proposed_block_diagram}
\end{figure*}
\section{SmaGla-FPAD: Face Presentation Attack Database for Smart Glass Capture}
\label{database}

Most existing face PAD benchmarks are acquired with fixed cameras or handheld devices and therefore do not fully capture the variability introduced by first person wearable acquisition. Smart glass capture is more challenging because it combines ego motion, viewpoint variation, background clutter, changing subject to sensor distance, and less controlled illumination. To study PAD under these conditions, we introduce \emph{SmaGla-FPAD}, a smart glass based face presentation attack database collected using Ray-Ban Meta Wayfarer Smart Glasses with a front facing RGB camera supporting 12 MP image capture and video recording up to 3K resolution. Figure~\ref{fig:data-acquisition} illustrates the acquisition process.

The database is collected in three settings, namely \emph{Indoor}, \emph{Outdoor}, and \emph{in-car}, chosen to reflect capture scenarios relevant to border control and mobile identity verification. The Indoor setting is acquired in a corridor under controlled lighting, whereas the Outdoor and in-car settings are recorded under natural daylight and therefore exhibit stronger variation in illumination, shadow, and background content. In  Indoor and Outdoor settings, subjects walk naturally towards the smart glass wearer from an initial distance of approximately 6~m, which introduces progressive changes in scale and pose. In the in-car setting, the subject remains seated while the smart glass device records from a distance of 1 to 2~m. To increase scene diversity, the in-car subset is collected across  10 vehicle types of different sizes.

\begin{table}[ht]
\centering
\small
\renewcommand{\arraystretch}{1.05}
\resizebox{\linewidth}{!}{
\begin{tabular}{|l|l|c|}
\hline
\textbf{Data Type} & \textbf{Condition} & \textbf{Total Samples} \\
\hline
Bona fide           & Indoor  & 75,000 \\
\hline
Bona fide           & Outdoor & 75,000 \\
\hline
Bona fide           & in-car  & 75,000 \\
\hline
Presentation attack & Indoor  & 24,000 \\
\hline
Presentation attack & Outdoor & 24,000 \\
\hline
Presentation attack & in-car  & 24,000 \\
\hline
\multicolumn{2}{|l|}{\textbf{Total Samples}} & \textbf{297,000} \\
\hline
\end{tabular}
}
\caption{Statistical summary of the proposed SmaGla-FPAD database.}
\label{tab:statistical-summary}
\end{table}

SmaGla-FPAD contains two classes, namely \emph{Bona fide} and \emph{Presentation attack}. In total, the database comprises 297,000 sample images, including 225,000 bona fide samples and 72,000 presentation attack samples. The bona fide subset contains recordings from 50 subjects aged between 18 and 40 years, including 18 male and 32 female participants. For the Indoor and Outdoor settings, each subject is recorded over multiple sessions spanning four months, with 5 to 8 videos per subject in each setting, and each sequence lasting between 30 and 50 seconds. In the in-car setting, recordings are acquired from both left and right sides of the vehicle. 

The presentation attack subset is generated using four custom built silicone masks, comprising two male and two female masks, worn by 15 subjects. Silicone masks are used because they represent a realistic three dimensional presentation attack instrument with high visual fidelity and substantial surface structure. To increase acquisition realism, accessories such as wigs, hoodies, and eyeglasses are incorporated during recording. Two attack conditions are considered: in the first, the attacker wears a silicone mask together with eyeglasses, a wig, and a hoodie; in the second, the attacker wears the same accessories except for eyeglasses. For each condition, 5 to 12 videos of 40 to 60 seconds are recorded over multiple sessions within 30 days. Table~\ref{tab:statistical-summary} summarises the dataset composition, and Figure~\ref{fig:sample-images} shows representative examples from the three settings.

\section{Proposed Method: DeCoupPAD-FAS++}
\label{sec:Prop}

We propose \emph{DeCoupPAD-FAS++}, a decoupled representation learning framework for face PAD under smart glass capture. The central idea is to learn a spoof discriminative representation while explicitly separating nuisance sensitive capture variation induced by first person wearable acquisition. This is achieved by decoupling the representation into two complementary branches: an \emph{attack content} branch for final PAD and a \emph{style domain} branch for nuisance modelling. The attack branch is further regularised by adversarial pseudo-domain suppression and prompt guided image-text alignment. An overview of the framework is shown in Fig.~\ref{fig:proposed_block_diagram}.

\subsection{Overall Architecture}

Given an input RGB face frame $x_i$, a frozen DINOv2 backbone extracts patch tokens $T_i \in \mathbb{R}^{N \times C}$, where $N$ and $C$ denote the number and dimension of patch tokens, respectively. Smart glass capture introduces substantial nuisance variation, including motion blur, illumination fluctuation, sharpness degradation, background clutter, and changes in subject to sensor distance. If not handled explicitly, such factors can become entangled with spoof cues and reduce generalisation. We therefore decompose the token representation into two branches with complementary roles.
The attack content branch learns a spoof discriminative embedding $z_i^{a}$ for final PAD, whereas the style domain branch learns a nuisance sensitive embedding $z_i^{d}$ that captures wearable capture variation. The attack branch aggregates informative token interactions through a lightweight Tiny Q-Former, followed by semantic adaptation and prompt gating to produce the final gated representation $\bar{z}_i^{a}$. In parallel, the style branch summarises the same token set through style sensitive statistics and projects them into $z_i^{d}$ under pseudo-domain supervision. To reduce nuisance leakage into the PAD representation, $\bar{z}_i^{a}$ is additionally constrained by adversarial pseudo-domain learning through a gradient reversal layer. During training, prompt guided image-text alignment with frozen OpenCLIP provides semantic regularisation. During inference, only the attack branch and PAD head are retained.
\vspace{-0.1in}
\subsection{Attack Content Branch}
The attack content branch transforms the patch tokens $T_i$ into a spoof discriminative representation. First, token interactions are selectively aggregated by a lightweight Tiny Q-Former:
\begin{equation}
Q_i = \phi_{\mathrm{qf}}(T_i),
\end{equation}
where $\phi_{\mathrm{qf}}(\cdot)$ denotes the Q-Former mapping. This design is motivated by the observation that spoof artefacts under smart glass capture are often local, spatially heterogeneous, and easily obscured by nuisance variation.

The aggregated tokens are then refined by a semantic adapter and projected into the attack embedding space:
\begin{equation}
\tilde{Q}_i = \phi_{\mathrm{sem}}(Q_i), \qquad
z_i^{a} = \phi_{\mathrm{att}}(\tilde{Q}_i),
\end{equation}
where $\phi_{\mathrm{sem}}(\cdot)$ and $\phi_{\mathrm{att}}(\cdot)$ denote the semantic adaptation and attack projection modules, respectively. A prompt gating module then recalibrates the resulting embedding:
\begin{equation}
\bar{z}_i^{a} = \mathrm{norm}(z_i^{a} \odot g_i),
\end{equation}
where $g_i$ is a learned gate vector, $\odot$ denotes element wise multiplication, and $\mathrm{norm}(\cdot)$ denotes feature normalisation. The final PAD score is computed solely from the gated attack representation:
\begin{equation}
\hat{y}_i = \sigma \big( h_{\mathrm{pad}}(\bar{z}_i^{a}) \big),
\end{equation}
where $h_{\mathrm{pad}}(\cdot)$ denotes the PAD head and $\sigma(\cdot)$ is the sigmoid function.

\subsection{Style Domain Branch}

The style domain branch is designed to model nuisance sensitive variation specific to smart glass acquisition. Starting from the same patch tokens $T_i$, this branch first computes style sensitive token statistics and applies MixStyle1D during training to improve robustness to appearance variation. The resulting features are then processed by a domain adapter and a domain MLP to produce the latent style representation
\begin{equation}
z_i^{d} = \phi_{\mathrm{sty}}(T_i),
\end{equation}
where $\phi_{\mathrm{sty}}(\cdot)$ denotes the overall style projection function implemented by the style statistics module, MixStyle1D, domain adapter, and domain MLP. In practice, this branch captures token level characteristics associated with blur, illumination, and related acquisition effects. The resulting representation is normalised and used for pseudo-domain supervision so that nuisance related factors are assigned to a dedicated latent space rather than contaminating the spoof representation.

\subsection{Pseudo-Domain Supervision and Adversarial Suppression}

Each training sample is assigned a pseudo-domain label $d_i \in \{1,\dots,K\}$ derived from simple image statistics, with $K = 9$ in our implementation. These labels act as lightweight proxies for nuisance conditions, such as brightness and blur, that are prominent under smart glass capture. The style embedding $z_i^{d}$ is passed to a domain head trained with pseudo-domain supervision so that this branch explicitly models nuisance sensitive variation.

In contrast, the gated attack embedding $\bar{z}_i^{a}$ is encouraged to suppress the same information. Specifically, $\bar{z}_i^{a}$ is passed through a gradient reversal layer followed by an adversarial domain head trained to predict $d_i$. Owing to gradient reversal, this objective discourages the attack branch from encoding pseudo-domain specific cues. The two branches therefore play complementary roles: the style branch captures nuisance variation, whereas the attack branch is encouraged to remain spoof discriminative while being less sensitive to capture specific factors.

\subsection{Prompt Guided Image-Text Alignment}

To further regularise the attack branch, we introduce prompt guided image-text alignment during training. A set of bona fide and attack prompt templates is first encoded by a frozen OpenCLIP text encoder. The resulting text embeddings are then mapped to the shared semantic space by a lightweight text projection layer. In parallel, the gated attack embedding $\bar{z}_i^{a}$ is projected into the same space by an image-to-text projection layer. Alignment is then enforced between the projected attack representation and the corresponding class-specific prompt embeddings. This branch regularises the attack representation using high level semantic guidance, thereby reducing overfitting to dataset specific low level artefacts and improving robustness under scenario variation. The image-text alignment module is used only during training and introduces no additional cost at inference.

\subsection{Training Objective}

The proposed framework is trained using a multi term objective that jointly optimises PAD discrimination, nuisance modelling, nuisance suppression, and semantic regularisation. Let $L_{\mathrm{pad}}$ denote the binary PAD loss, $L_{\mathrm{dom}}$ the pseudo-domain classification loss on the style branch, $L_{\mathrm{adv}}$ the adversarial pseudo-domain loss applied through the gradient reversal layer on $\bar{z}_i^{a}$, $L_{\mathrm{itc}}$ and $L_{\mathrm{ptm}}$ the prompt guided image-text alignment losses, $L_{\mathrm{orth}}$ the orthogonality regulariser between $z_i^{a}$ and $z_i^{d}$, and $L_{\mathrm{style}}$ the style regularisation term. The overall loss is defined as
\begin{equation}
\begin{aligned}
L_{\mathrm{total}} =\;& L_{\mathrm{pad}}
+ \lambda_{\mathrm{dom}} L_{\mathrm{dom}}
+ \lambda_{\mathrm{adv}} L_{\mathrm{adv}}
+ \lambda_{\mathrm{itc}} L_{\mathrm{itc}} \\
&+ \lambda_{\mathrm{ptm}} L_{\mathrm{ptm}}
+ \lambda_{\mathrm{orth}} L_{\mathrm{orth}}
+ \lambda_{\mathrm{style}} L_{\mathrm{style}} .
\end{aligned}
\end{equation}
where $\lambda_{\mathrm{dom}}, \lambda_{\mathrm{adv}}, \lambda_{\mathrm{itc}}, \lambda_{\mathrm{ptm}}, \lambda_{\mathrm{orth}}$, and $\lambda_{\mathrm{style}}$ are scalar weighting coefficients.

Here, $L_{\mathrm{pad}}$ drives bona fide versus attack discrimination, $L_{\mathrm{dom}}$ encourages $z_i^{d}$ to encode nuisance related variation, and $L_{\mathrm{adv}}$ suppresses the same information in $\bar{z}_i^{a}$. The orthogonality term further separates the attack and style subspaces, while $L_{\mathrm{itc}}$ and $L_{\mathrm{ptm}}$ provide semantic regularisation for the attack branch through prompt guided alignment.

\subsection{Inference Stage}
At inference, an input frame $x_i$ is passed through the frozen DINOv2 backbone and the attack content branch to obtain the gated representation $\bar{z}_i^{a}$, from which the final PAD score is computed as
\begin{equation}
\hat{y}_i = \sigma \big( h_{\mathrm{pad}}(\bar{z}_i^{a}) \big).
\end{equation}
The style domain branch and all auxiliary modules used for pseudo-domain supervision, adversarial suppression, and prompt guided image-text alignment are used only during training and are discarded at test time, ensuring a simple and efficient inference pipeline.

\section{Experiments and Results}
\label{sec:Exp}

We first describe the experimental setup and evaluation protocols, and then present the results for in domain and cross scenario presentation attack detection.

\subsection{Experimental Setup and Protocols}

We evaluate the proposed DeCoupPAD-FAS++ under two smart glass acquisition settings, namely \emph{Car} and \emph{Moving}. The Moving setting is further divided into indoor and outdoor subsets and therefore introduces larger variability in pose, motion, blur, illumination, and background conditions than the Car setting. Based on these two acquisition settings, we define four experimental protocols to assess both in domain presentation attack detection and cross scenario generalisation.

For all experiments, the data are split using subject disjoint partitions for both bona fide samples and silicone mask attacks. The bona fide class contains 50 unique subjects, of which 35 are used for training, 5 for validation, and the remaining 10 for testing. The attack class contains four distinct silicone masks. Among these, two masks are used for training, one for validation, and one for testing. The same splitting strategy is repeated across three independent runs, and the final performance is reported as the mean together with the corresponding 95\% confidence interval.

Evaluation is conducted exclusively on the test set. We report BPCER at the operating point corresponding to APCER equal to $1\%$, the resulting APCER measured on the test set, and the corresponding ACER. In all cases, the decision threshold is estimated on the validation set and then fixed for test evaluation. This protocol allows us to analyse not only the in domain effectiveness of the proposed method but also its ability to retain spoof discriminative cues under scenario change. 
\vspace{-0.2in}
\paragraph{Experiment E1: In domain PAD in the Car scenario.}
The first protocol evaluates in domain PAD in the Car scenario using still face frames. Training and validation are performed exclusively on Car data, where bona fide samples and silicone mask attacks are both drawn from the Car scenario according to the subject disjoint partitions. Testing is then carried out on the corresponding disjoint Car test subjects and test attacks. Since both optimisation and evaluation are performed within the same acquisition setting, E1 serves as the most controlled protocol and provides the in domain reference performance of the proposed method.
\vspace{-0.2in}
\paragraph{Experiment E2: In domain PAD in the Moving scenario.}
The second protocol evaluates in domain PAD in the Moving scenario. This setting is more challenging than the Car scenario because it includes subject motion and stronger environmental variation. Training and validation use bona fide samples and silicone mask attacks captured under the Moving scenario, following the same subject disjoint partitions. Testing is performed on the corresponding disjoint Moving test subjects and test attacks. This protocol measures the ability of the proposed method to detect attacks under more variable acquisition conditions.

To make the effect of environmental variation explicit, we additionally report two test only subsets using the same trained model. In \textbf{E2a}, evaluation is restricted to the \emph{Moving Indoor} subset, whereas in \textbf{E2b}, evaluation is restricted to the \emph{Moving Outdoor} subset. These two analyses do not introduce additional training, but instead isolate the influence of indoor and outdoor capture conditions on the final PAD performance.
\vspace{-0.2in}
\paragraph{Experiment E3: Cross scenario generalisation from Moving to Car.}
The third protocol evaluates cross scenario generalisation from the Moving scenario to the Car scenario. Training and validation are performed using bona fide samples and silicone mask attacks from the Moving scenario, including both indoor and outdoor data, according to the subject disjoint partitions. Testing is then carried out on the corresponding disjoint Car test subjects and test attacks. This protocol is more demanding than the in domain setting because the training and test data arise from different acquisition conditions. It therefore measures whether the proposed method learns spoof related cues that remain effective across scenario change rather than relying on scenario specific characteristics.
\vspace{-0.2in}
\paragraph{Experiment E4: Cross scenario generalisation from Car to Moving.}
The fourth protocol evaluates cross scenario generalisation in the reverse direction, namely from the Car scenario to the Moving scenario. Training and validation are performed using bona fide samples and silicone mask attacks from the Car scenario, following the same subject disjoint partitions, whereas testing is carried out on the corresponding disjoint Moving test subjects and test attacks. For completeness, we report results on the combined Moving test set as well as separately on the indoor and outdoor subsets using the same trained model. This protocol assesses whether a model trained under relatively constrained Car conditions remains effective when transferred to the more variable Moving scenario.

\subsection{Results and Discussion}

\begin{table}[t]
\centering
\caption{Results for E1 under the Car scenario. Performance is reported as mean with 95\% confidence interval over three runs. Lower values are better.}
\label{tab:e1_results}
\setlength{\tabcolsep}{4pt}
\renewcommand{\arraystretch}{1.12}
\resizebox{\columnwidth}{!}{%
\begin{tabular}{|l|c|c|c|}
\hline
\textbf{PAD Method} & \textbf{BPCER@APCER=1\%} & \textbf{APCER@thr=1\%} & \textbf{ACER} \\
\hline
CFPL   \cite{liu2024cfpl}          & 0.59 [0.00, 99.11] & 1.44 [0.07, 13.26] & 1.01 [0.08, 4.51] \\ \hline
InstructFLIP \cite{lin2025instructflip}     & 0.39 [0.00, 99.99] & 0.44 [0.02, 4.32]  & 0.41 [0.04, 2.95] \\ \hline
SAFAS \cite{SAFAS}          & 0.00 [0.00, 0.00]  & 16.05 [0.00, 99.01] & 8.02 [0.01, 97.21] \\ \hline
TTSP \cite{zhou2024test}           & 2.16 [0.00, 100.00] & 4.16 [0.00, 92.20] & 3.39 [0.22, 24.17] \\ \hline
MEFAS \cite{MEFAS}           & 67.29 [33.14, 90.31] & 1.25 [0.01, 29.44] & 34.27 [22.01, 48.78] \\ \hline
MTFAS \cite{MTFAS}           & 49.95 [0.88, 99.40] & 0.00 [0.00, 0.00]  & 24.97 [2.33, 77.10] \\ \hline
\textbf{Proposed Method} & \textbf{0.00 [0.00, 0.00]} & \textbf{0.23 [0.01, 2.56]} & \textbf{0.11 [0.00, 2.13]} \\
\hline
\end{tabular}%
}
\end{table}
Table~\ref{tab:e1_results} reports the results of Experiment E1 in the Car scenario. All compared methods were implemented using publicly available open source code where available, and otherwise reimplemented, and evaluated under the same subject disjoint splits and experimental protocol as the proposed method.  The proposed method achieves the best overall performance, with the lowest ACER of 0.11\% [0.00, 2.13] and the lowest APCER at the validation derived operating point, namely 0.23\% [0.01, 2.56]. This indicates that the proposed model provides the most favourable balance between attack rejection and bona fide acceptance under matched training and test conditions.

The compared methods show clear trade offs. SAFAS attains zero BPCER at APCER = 1\%, but exhibits a much larger APCER of 16.05\%, indicating weak attack rejection. MTFAS achieves zero APCER, but only at the cost of a very large BPCER of 49.95\%. MEFAS shows a similar tendency, with a relatively low APCER of 1.25\% but a very large BPCER of 67.29\%, leading to poor overall ACER. CFPL, InstructFLIP, and TTSP provide more balanced behaviour than these extreme cases, yet all remain inferior to the proposed method. Overall, the E1 results establish that the proposed representation provides the strongest in domain performance in the controlled Car scenario.

\begin{table}[htp]
\centering
\caption{Results for E2 under the Moving scenario. Performance is reported as mean with 95\% confidence interval over three runs for the Indoor, Outdoor, and Both subsets. Lower values are better. Best results in each subset are shown in bold.}
\label{tab:e2_results}
\setlength{\tabcolsep}{6pt}
\renewcommand{\arraystretch}{1.15}
\small
\resizebox{0.95\columnwidth}{!}{%
\begin{tabular}{|l|c|c|c|}
\hline
\textbf{PAD Method} & \textbf{BPCER@APCER=1\%} & \textbf{APCER@thr=1\%} & \textbf{ACER} \\
\hline
\multicolumn{4}{|c|}{\textbf{Indoor}} \\
\hline
CFPL  \cite{liu2024cfpl}           & 27.12 [0.76, 84.76]          & 2.49 [0.00, 100.00]         & 14.80 [1.10, 53.00]          \\
\hline
InstructFLIP \cite{lin2025instructflip}     & 28.11 [0.07, 98.04]          & 4.94 [0.00, 100.00]         & 16.53 [2.98, 50.81]          \\
\hline
SAFAS \cite{SAFAS}           & 24.53 [0.32, 92.96]          & 4.00 [0.00, 100.00]         & 14.27 [2.19, 48.46]          \\
\hline
TTSP  \cite{zhou2024test}            & 51.40 [0.21, 99.86]          & 4.31 [0.00, 100.00]         & 27.89 [2.42, 81.44]          \\
\hline
MEFAS   \cite{MEFAS}         & 99.17 [96.03, 99.87]         & 1.13 [0.88, 1.44]           & 50.15 [49.65, 50.64]         \\
\hline
MTFAS  \cite{MTFAS}          & 99.96 [1.84, 100.00]         & \textbf{0.01 [0.00, 88.05]} & 49.98 [49.88, 50.08]         \\
\hline
Proposed Method & \textbf{20.59 [0.07, 95.71]} & 5.15 [0.00, 100.00]        & \textbf{12.87 [2.43, 41.05]} \\
\hline
\multicolumn{4}{|c|}{\textbf{Outdoor}} \\
\hline
CFPL \cite{liu2024cfpl}           & 42.57 [2.98, 94.48]          & 2.49 [0.00, 100.00]         & 22.53 [3.87, 63.10]          \\
\hline
InstructFLIP \cite{lin2025instructflip}    & 34.10 [0.30, 97.69]          & 4.94 [0.00, 100.00]         & 19.52 [2.23, 65.17]          \\
\hline
SAFAS  \cite{SAFAS}          & 32.07 [0.27, 97.40]          & 4.00 [0.00, 100.00]         & 18.04 [1.51, 67.51]          \\
\hline
TTSP   \cite{zhou2024test}           & 65.74 [1.19, 99.83]          & 4.31 [0.00, 100.00]         & 35.02 [5.06, 81.85]          \\
\hline
MEFAS \cite{MEFAS}           & 94.64 [91.05, 96.94]         & 1.13 [0.88, 1.44]           & 47.89 [46.28, 49.49]         \\
\hline
MTFAS  \cite{MTFAS}         & 99.81 [0.18, 100.00]         & \textbf{0.01 [0.00, 88.05]} & 49.93 [49.60, 50.26]         \\
\hline
Proposed Method & \textbf{29.98 [0.11, 98.29]} & 5.16 [0.00, 100.00]        & \textbf{17.57 [0.87, 74.01]} \\
\hline
\multicolumn{4}{|c|}{\textbf{Both}} \\
\hline
CFPL   \cite{liu2024cfpl}          & 31.84 [2.31, 87.57]          & 2.49 [0.00, 100.00]         & 17.16 [2.47, 56.29]          \\
\hline
InstructFLIP \cite{lin2025instructflip}     & 30.81 [0.30, 96.43]          & 4.94 [0.00, 100.00]         & 17.88 [3.06, 54.50]          \\
\hline
SAFAS  \cite{SAFAS}          & 27.81 [0.36, 94.75]          & 4.00 [0.00, 100.00]         & 15.91 [1.90, 56.91]          \\
\hline
TTSP  \cite{zhou2024test}            & 57.84 [1.03, 99.55]          & 4.31 [0.00, 100.00]         & 31.07 [4.16, 79.01]          \\
\hline
MEFAS \cite{MEFAS}           & 97.08 [94.76, 98.43]         & 1.13 [0.88, 1.44]           & 49.10 [48.28, 49.91]         \\
\hline
MTFAS \cite{MTFAS}           & 99.99 [0.47, 100.00]         & \textbf{0.01 [0.00, 88.05]} & 49.96 [49.76, 50.16]         \\
\hline
Proposed Method & \textbf{24.71 [0.18, 95.15]} & 5.16 [0.00, 100.00]        & \textbf{14.93 [1.67, 55.72]} \\
\hline
\end{tabular}
}
\vspace{-.18in}
\end{table}

The quantitative results of Experiment E2 are summarised in Table~\ref{tab:e2_results}. Compared with E1, all methods exhibit degraded performance, confirming that the Moving scenario is substantially more challenging due to motion, pose change, and environmental variability. On the combined test set, which provides the most representative evaluation for this protocol, the proposed method achieves the lowest ACER of 14.93\% [1.67, 55.72] together with the lowest BPCER at APCER = 1\%, namely 24.71\% [0.18, 95.15]. On the Indoor subset, the proposed method attains the lowest BPCER and the lowest ACER, while on the Outdoor subset it also achieves both the lowest BPCER and the lowest ACER. These observations are important because the Outdoor subset constitutes the more challenging condition within the Moving scenario.

Several compared methods obtain smaller APCER values than the proposed method, particularly MTFAS and MEFAS. However, this comes at the cost of substantially increased bona fide rejection. MTFAS exhibits extremely large BPCER across all three subsets, while MEFAS shows similarly poor BPCER despite its relatively low APCER. As a result, both methods yield much larger ACER than the proposed method. CFPL, InstructFLIP, and SAFAS provide more balanced behaviour than these extreme cases, yet they remain inferior to the proposed method on the most representative combined setting. Overall, these results indicate that the proposed method provides the best balance between attack rejection and bona fide acceptance under the more challenging Moving scenario, both across individual environmental conditions and on the combined test set.

\begin{table}[t]
\centering
\caption{Results for E3, cross scenario generalisation from Moving to Car. Performance is reported as mean with 95\% confidence interval over three runs. Lower values are better.}
\label{tab:e3_results}
\setlength{\tabcolsep}{5pt}
\renewcommand{\arraystretch}{1.12}
\footnotesize
\resizebox{\columnwidth}{!}{%
\begin{tabular}{|l|c|c|c|}
\hline
\textbf{PAD Method} & \textbf{BPCER@APCER=1\%} & \textbf{APCER@thr=1\%} & \textbf{ACER} \\
\hline
CFPL  \cite{liu2024cfpl}           & 59.25 [0.27, 99.85]           & 0.00 [0.00, 0.00] & 29.63 [0.36, 95.31]          \\
\hline
InstructFLIP \cite{lin2025instructflip}     & 62.98 [0.15, 99.95]           & 0.00 [0.00, 0.00] & 31.49 [0.30, 96.53]          \\
\hline
SAFAS \cite{SAFAS}           & 67.50 [0.02, 100.00]          & 0.06 [0.00, 99.31] & 33.78 [0.50, 96.01]         \\
\hline
TTSP \cite{zhou2024test}             & 67.03 [0.00, 100.00]          & 0.00 [0.00, 0.00] & 33.51 [0.37, 96.64]          \\
\hline
MEFAS \cite{MEFAS}           & 99.99 [19.89, 100.00]       & 0.00 [0.00, 0.00] & 49.99 [49.97, 50.01]         \\
\hline
MTFAS  \cite{MTFAS}          & 100.00 [100.00, 100.00]       & 0.00 [0.00, 0.00] & 50.00 [50.00, 50.00]         \\
\hline
Proposed Method & \textbf{38.37 [0.00, 99.91]} & 0.04 [0.00, 98.53] & \textbf{19.20 [0.00, 98.99]} \\
\hline
\end{tabular}%
}
\vspace{-.2in}
\end{table}

Table~\ref{tab:e3_results} presents the results of Experiment E3, which evaluates cross scenario generalisation from Moving to Car. This is a challenging setting because the model must transfer spoof related cues learned under motion and environmental variability to a different and more constrained acquisition condition. Under this protocol, the proposed method achieves the lowest BPCER at APCER = 1\%, 38.37\% [0.00, 99.91], and the lowest ACER, 19.20\% [0.00, 98.99]. This indicates that the proposed method transfers more effectively across scenarios than the compared baselines.

Although several compared methods achieve APCER values of 0.00\%, this occurs together with very large BPCER values, indicating overly conservative behaviour that rejects a substantial proportion of bona fide samples. This tendency is especially clear for MEFAS and MTFAS. In contrast, the proposed method maintains APCER close to zero while substantially reducing BPCER, resulting in the best overall trade off between attack rejection and bona fide acceptance. These results suggest that the proposed representation captures spoof related cues that generalise better across acquisition conditions rather than relying primarily on scenario specific artefacts.

\begin{table}[htp]
\centering
\caption{Results for E4, cross scenario generalisation from Car to Moving. Performance is reported as mean with 95\% confidence interval over three runs for the Indoor, Outdoor, and Both subsets. Lower values are better. Best results in each subset are shown in bold.}
\label{tab:e4_results}
\setlength{\tabcolsep}{5pt}
\renewcommand{\arraystretch}{1.12}
\resizebox{0.95\columnwidth}{!}{%
\begin{tabular}{|l|c|c|c|}
\hline
\textbf{PAD Method} & \textbf{BPCER@APCER=1\%} & \textbf{APCER@thr=1\%} & \textbf{ACER} \\
\hline
\multicolumn{4}{|c|}{\textbf{Indoor}} \\
\hline
CFPL \cite{liu2024cfpl}            & 2.12 [0.00, 66.28]          & 53.72 [26.57, 79.02]         & 27.92 [13.91, 47.28]         \\
\hline
InstructFLIP \cite{lin2025instructflip}    & 1.56 [0.00, 99.99]          & 33.51 [10.31, 67.08]         & 17.53 [8.14, 32.66]          \\
\hline
SAFAS  \cite{SAFAS}          & 0.01 [0.00, 99.54]          & 82.80 [16.61, 99.62]         & 41.41 [23.04, 62.17]         \\
\hline
TTSP \cite{zhou2024test}             & 0.33 [0.00, 99.99]          & 88.33 [18.13, 99.90]         & 44.33 [26.82, 63.16]         \\
\hline
MEFAS   \cite{MEFAS}         & 26.44 [0.40, 93.54]         & 22.42 [1.91, 74.76]         & 24.43 [4.50, 64.84]         \\
\hline
MTFAS \cite{MTFAS}           & 6.75 [0.00, 99.29]          & 100.00 [100.00, 100.00]      & 53.37 [36.76, 88.16]         \\
\hline
Proposed Method & 2.84 [0.00, 86.31]          & \textbf{20.52 [6.76, 45.53]} & \textbf{11.68 [5.90, 21.08]} \\
\hline
\multicolumn{4}{|c|}{\textbf{Outdoor}} \\
\hline
CFPL   \cite{liu2024cfpl}          & 0.73 [0.02, 7.61]           & 53.72 [26.57, 79.02]         & 27.22 [15.00, 43.56]         \\
\hline
InstructFLIP \cite{lin2025instructflip}   & 9.51 [0.00, 100.00]         & 33.52 [10.31, 67.08]         & 21.51 [15.68, 28.58]         \\
\hline
SAFAS  \cite{SAFAS}          & \textbf{0.00 [0.00, 0.00]} & 82.80 [16.61, 99.62]         & 41.40 [23.03, 62.17]         \\
\hline
TTSP  \cite{zhou2024test}            & 0.55 [0.03, 4.88]           & 88.33 [18.13, 99.90]         & 44.40 [26.87, 63.30]         \\
\hline
MEFAS  \cite{MEFAS}          & 18.71 [0.09, 93.15]         & 22.42 [1.91, 74.76]          & 20.57 [3.04, 62.59]         \\
\hline
MTFAS \cite{MTFAS}           & 18.65 [0.00, 97.91]         & 100.00 [100.00, 100.00]      & 68.65 [55.76, 70.16]         \\
\hline
Proposed Method & 1.09 [0.00, 42.45]          & \textbf{20.52 [6.76, 45.53]} & \textbf{10.80 [4.87, 21.27]} \\
\hline
\multicolumn{4}{|c|}{\textbf{Both}} \\
\hline
CFPL   \cite{liu2024cfpl}          & 1.33 [0.01, 25.57]          & 53.72 [26.57, 79.02]         & 27.52 [14.52, 45.13]         \\
\hline
InstructFLIP \cite{lin2025instructflip}     & 6.18 [0.00, 100.00]         & 33.51 [10.31, 67.08]         & 19.84 [17.12, 22.84]         \\
\hline
SAFAS  \cite{SAFAS}          & \textbf{0.00 [0.00, 97.91]} & 82.80 [16.61, 99.62]        & 41.40 [23.03, 62.17]         \\
\hline
TTSP \cite{zhou2024test}             & 0.43 [0.14, 1.10]           & 88.33 [18.13, 99.90]         & 44.38 [26.86, 63.20]         \\
\hline
MEFAS \cite{MEFAS}          & 22.47 [0.28, 91.82]         & 22.42 [1.91, 74.76]          & 22.45 [4.03, 62.07]         \\
\hline
MTFAS \cite{MTFAS}           & 12.70 [0.00, 92.91]         & 100.00 [100.00, 100.00]      & 61.01 [52.76, 80.16]         \\
\hline
Proposed Method & 1.99 [0.00, 73.41]          & \textbf{20.52 [6.76, 45.53]} & \textbf{11.25 [5.56, 20.66]} \\
\hline
\end{tabular}%
}
\vspace{-.2in}
\end{table}
The results of Experiment E4 are shown in Table~\ref{tab:e4_results}, where the model is trained on the Car scenario and evaluated on the Moving scenario. This protocol is complementary to E3 and is particularly important because cross scenario transfer is not necessarily symmetric. The proposed method achieves the lowest ACER across all three evaluation subsets, with values of 11.68\% [5.90, 21.08] for Indoor, 10.80\% [4.87, 21.27] for Outdoor, and 11.25\% [5.56, 20.66] for Both. It also attains the lowest APCER in all three cases, indicating stronger suppression of attack acceptance after transfer from the more constrained Car scenario to the more variable Moving scenario.

Several compared methods achieve lower BPCER than the proposed method, but this occurs together with substantially larger APCER, which leads to inferior ACER. This tendency is particularly pronounced for SAFAS, TTSP, and MTFAS, all of which favour bona fide acceptance at the expense of weak attack rejection. MEFAS provides a more balanced operating point than these more extreme cases, yet it remains consistently inferior to the proposed method across Indoor, Outdoor, and Both evaluation subsets. In contrast, the proposed method provides the most favourable overall trade off between attack rejection and bona fide acceptance under cross scenario transfer. In summary, the E4 results show that the proposed representation transfers effectively from Car to Moving while maintaining the strongest overall operating point under the stronger nuisance variation of the Moving scenario.

Across all four protocols, a consistent pattern emerges. In the controlled in domain setting of E1, the proposed method achieves the best overall performance. In the more challenging in domain setting of E2, it remains the most balanced method across varying environmental conditions. More importantly, in both cross scenario protocols, E3 and E4, the proposed model achieves the most favourable trade off between attack rejection and bona fide acceptance. These findings support the central design objective of the proposed method, namely to learn spoof discriminative representations that are less entangled with nuisance sensitive capture variation and therefore generalise more effectively across changing acquisition conditions.

\subsection{Ablation Study}
Table~\ref{tab:ablation_e3} reports the ablation study on the E3 protocol. The full model achieves the lowest BPCER and ACER, indicating that the complete formulation provides the most effective balance between attack rejection and bona fide acceptance. The CLS baseline and patch mean baseline perform substantially worse, showing that generic token aggregation is insufficient for robust cross-scenario PAD. Using only the attack branch already improves performance, whereas adding the style branch without proper disentanglement degrades it, confirming that spoof related and nuisance related factors must be separated rather than simply combined.
\begin{table}[htp]
\centering
\caption{Ablation study on the E3 protocol. Results are reported as mean over three independent runs. Lower values are better.}
\label{tab:ablation_e3}
\setlength{\tabcolsep}{4pt}
\renewcommand{\arraystretch}{1.12}
\footnotesize
\resizebox{\columnwidth}{!}{%
\begin{tabular}{|l|c|c|c|}
\hline
\textbf{Ablation} & \textbf{BPCER@APCER=1\%} & \textbf{APCER@thr=1\%} & \textbf{ACER\%} \\
\hline
CLS baseline                          & 74.64 & 0.000 & 37.32 \\
\hline
Patch mean baseline                   & 64.88 & 0.000 & 32.44 \\
\hline
Attack only                           & 54.87 & 0.000 & 27.44 \\
\hline
Attack + style, no disentanglement    & 62.56 & 0.011 & 31.28 \\
\hline
Full w/o $L_{\mathrm{orth}}$          & 63.71 & 0.000 & 31.86 \\
\hline
Full w/o pseudo-domain losses         & 57.89 & 0.000 & 28.95 \\
\hline
Full w/o prompt alignment             & 62.30 & 0.022 & 31.16 \\
\hline
Full + ITC only                       & 55.38 & 0.022 & 27.70 \\
\hline
Full + PTM only                       & 56.65 & 0.133 & 28.40 \\
\hline
\textbf{Full model}                   & \textbf{38.37} & \textbf{0.044} & \textbf{19.19} \\
\hline
\end{tabular}%
}
\vspace{-.2in}
\end{table}
A similar trend is observed in the remaining variants. Removing the orthogonality constraint, pseudo-domain losses, or prompt alignment each increases ACER relative to the full model, while using only ITC or only PTM remains inferior to the complete formulation. Overall, the gain comes from the combined effect of disentanglement, pseudo-domain supervision, and prompt based alignment rather than from any single component alone.
\section{Conclusion}
\label{sec:conc}
We presented \emph{SmaGla-FPAD}, a smart glass based face presentation attack database, and \emph{DeCoupPAD-FAS++}, a decoupled framework for face PAD under first person wearable capture. The proposed method separates spoof-related evidence from nuisance-sensitive capture variation and achieves a strong balance between attack rejection and bona fide acceptance across both in-domain and cross-scenario protocols. The ablation study further confirms that disentanglement, pseudo-domain supervision, and prompt guided regularisation are all important for robust generalisation under smart glass capture. A limitation of the present work is that the evaluation is conducted on a single smart glass dataset with a limited set of silicone mask attacks. Future work will extend the framework to broader attack types, additional presentation attack instruments and wearable capture devices, and cross-dataset evaluation.
{\small
\bibliographystyle{ieee}
\bibliography{2DFacePAD}
}

\end{document}